\documentclass{article}

\usepackage{microtype}
\usepackage{graphicx}
\usepackage{booktabs}
\usepackage{subcaption}
\usepackage{hyperref}
\usepackage{multirow}

\usepackage[accepted]{icml2025}

\usepackage{amsmath}
\usepackage{amssymb}
\usepackage{mathtools}
\usepackage{amsthm}

\usepackage[capitalize,noabbrev]{cleveref}

\theoremstyle{plain}

\theoremstyle{definition}

\theoremstyle{remark}

\icmltitlerunning{Can Image-To-Video Models Simulate Pedestrian Dynamics?}

\begin{document}

\twocolumn[
\icmltitle{Can Image-To-Video Models Simulate Pedestrian Dynamics?}

\icmlsetsymbol{equal}{*}

\begin{icmlauthorlist}
\icmlauthor{Aaron Appelle}{duke}
\icmlauthor{Jerome P. Lynch}{duke}
\end{icmlauthorlist}

\icmlaffiliation{duke}{Duke University, Durham, NC, USA}

\icmlcorrespondingauthor{Aaron Appelle}{aaron.appelle@duke.edu}

\icmlkeywords{video generation, image-to-video, pedestrian behavior, pedestrian dynamics, trajectory prediction, multi-agent}

\vskip 0.3in
]

\printAffiliationsAndNotice{} 

\begin{abstract}
Recent high-performing image-to-video (I2V) models based on variants of the diffusion transformer (DiT) have displayed remarkable inherent world-modeling capabilities by virtue of training on large scale video datasets.
We investigate whether these models can generate realistic pedestrian movement patterns in crowded public scenes.
Our framework conditions I2V models on keyframes extracted from pedestrian trajectory benchmarks, then evaluates their trajectory prediction performance using quantitative measures of pedestrian dynamics.
\end{abstract}

\section{Introduction}
Realistic modeling and synthesis of pedestrian behavior is paramount for diverse applications including autonomous driving \cite{jiang_motiondiffuser_2023}, urban planning \cite{zhou_crowd_2010,al-kodmany_crowd_2013}, interactive robotics \cite{nocentini_survey_2019,jahanmahin_human-robot_2022}, and the generation of realistic virtual content \cite{reynolds1999steering,ARMANTO2025100875}.
Early computational approaches, often based on physics-inspired rules and forces \cite{helbing1995social,van2008reciprocal,bradley1993proposed}, provided foundational insights but struggled to capture the full complexity of real-world human interactions \cite{bae_continuous_2025}.
Deep learning methodologies marked a significant leap forward, employing LSTMs for path forecasting \cite{alahi_social_2016}, GANs for socially aware trajectory synthesis \cite{gupta_social_2018}, and graph-based networks to model inter-agent dynamics \cite{mohamed_social-stgcnn_2020, salzmann_trajectron_2020,shi_sgcn_2021}. More recently, generative frameworks like diffusion models \cite{gu_stochastic_2022,jiang_motiondiffuser_2023} and transformers \cite{yuan_agentformer_2021,chib_ms-tip_2024} have excelled at synthesizing complex multi-agent behaviors and interactions \cite{ribeiro-gomes_motiongpt_2024}. 
However, most models tackle the specific task of predicting a finite sequence of future steps given historical inputs, struggling at scene population and zero-shot scene transfer \cite{trajclip_2024}.

\begin{figure}[t!]
	\centering
	\includegraphics[width=\linewidth]{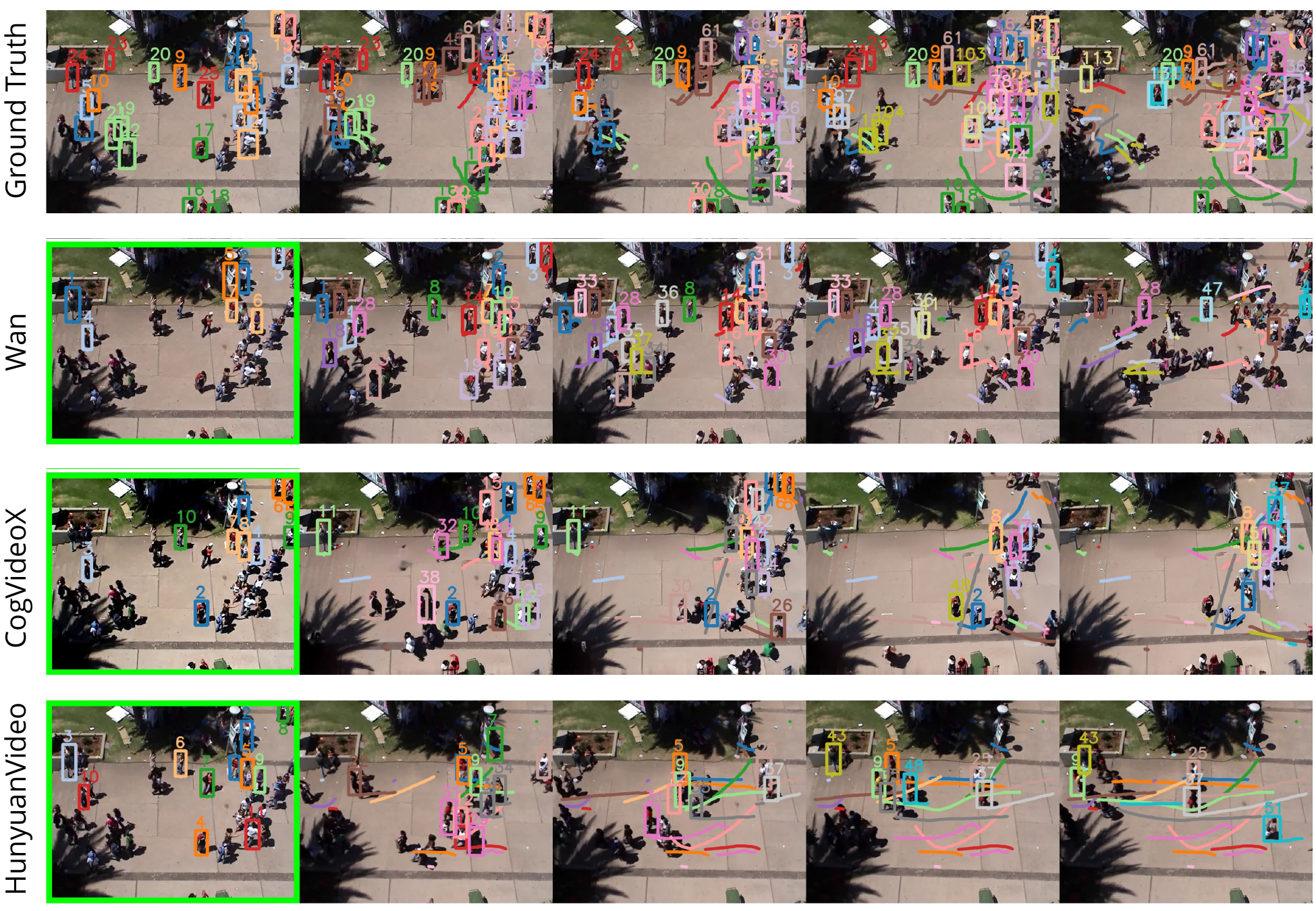}
    \caption{A 5-second excerpt from the UNIV scene of the ETH/UCY benchmark showing sample video generations using first-frame conditioning. Green borders indicate conditioning frames, and all frames show tracklets of detected pedestrians. The extracted frames are equally spaced at 1.25 seconds.}
	\label{fig:frames}
\end{figure}

Large-scale video generation models \cite{yang_cogvideox_2025,wan_wan_2025,kong_hunyuanvideo_2025,ju_fulldit_2025,chen_videocrafter2_2024,zheng2024open} have demonstrated an implicit capacity for world understanding \cite{brooks_video_2024}, spurring research into their potential as general-purpose world simulators capable of rendering visual appearance, physical phenomena, and agent interactions. Initial studies leveraging video generation for physics-based tasks, such as rigid-body dynamics \cite{leonardis_physgen_2025,montanaro_motioncraft_nodate} and object interactions \cite{zhang_physics-based_nodate}, have shown promise.

While pedestrian behavior exhibits physics-like properties, it also incorporates human decision-making that creates emergent social phenomena.
This positions crowd dynamics as an ideal benchmark for testing whether video generation models can simulate more complex human systems.
Open-source image-to-video (I2V) models achieve high visual fidelity and score increasingly well on general video quality benchmarks \cite{huang_vbench_2024,zheng_vbench-20_2025}. However, the evaluation metrics are geared towards videos that include one or a few identifiable humans, and do not directly apply to wide-angle scenes with many distant people that may lack identifiable faces, clothing, or gestures. The physical correctness and behavioral plausibility of multi-agent social navigation produced by video generation models has not yet been systematically evaluated.

In this work, we propose a benchmarking framework to evaluate how well I2V models simulate realistic multi-person behavior. We condition I2V models with images extracted from the ETH/UCY datasets \cite{Lerner_ETH,Pellegrini_UCY} to generate large sets of synthetic videos. Then, we use an off-the-shelf multi-object tracker (MOT) to track pedestrians and evaluate a suite of pedestrian dynamics metrics against the ground truth ETH/UCY data.
We observe that I2V models reproduce general motion patterns such as walking speed and direction but often fail at modeling agent-to-agent interactions, leading to collisions, disappearances, or unrealistic behaviors. 

\section{Method}
\subsection{I2V Inference}
The first step in our framework is to compute a large dataset of synthetic videos simulating the four public scenes present in the ETH/UCY datasets. The datasets depict a total of 1536 pedestrians in crowded settings like plazas, crossing points, and sidewalks \cite{gupta_social_2018}. The native resolution of the source videos is 480p for the ETH scene and 576p for the other scenes (HOTEL, UNIV, ZARA).
We slide through the video footage of each of the ETH/UCY scenes to extract start frames of 5-second video clips.
We selected three state-of-the-art I2V models: Wan2.1-I2V-14B-480P \cite{wan_wan_2025}, CogVideoX1.5-5B-I2V \cite{yang_cogvideox_2025}, and HunyuanVideo-I2V \cite{kong_hunyuanvideo_2025}.
To match their default training settings for best performance, we set Wan2.1 and CogVideoX to generate 480p videos of 81 frames at 16 fps, and HunyuanVideo to produce 129-frame clips at 25 fps, resulting in 5-second clips.
We provide a constant text prompt for all generations and all scenes: ``\textit{A stationary overhead view of pedestrian movement.}''

\subsection{Postprocessing}
After performing inference over the set of start images, we use FairMOT \cite{zhang_fairmot_2021} as an off-the-shelf pedestrian detection and tracking model.
The tracker produces frame-by-frame pixel coordinates of bounding boxes with a unique identifier for each tracked subject.
We estimate the contact of each person with the ground as the bottom midpoint of the bounding box. Then we use the provided homography matrices to convert pixel coordinates to the world coordinate system.

The original ETH/UCY datasets were manually annotated, providing complete coverage of all pedestrians that enter and exit the scenes. In contrast, we observe that FairMOT frequently misses pedestrians in the synthetic videos. This likely stems from two factors:
(1) a distribution mismatch between the human pixel representations generated by diffusion models and the real-world visual training data of FairMOT;
and (2) FairMOT's inherent limitations, as it occasionally misses detections even on real videos. The distribution mismatch creates a coupling between the aesthetic quality of I2V models and pedestrian detection performance: higher quality I2V models produce more realistic human visualizations that are successfully detected in postprocessing.
To correct for this factor, we perform multiple model inferences until we have a total of at least $N=1000$ pedestrian detections per scene.
Furthermore, the nearly birds-eye view camera angles in ETH/UCY datasets exacerbate the detection difficulties, especially at 480p resolution. To eliminate bias from using different detection methods on real versus synthetic data, we re-generate the ground truth tracking data on the ETH/UCY videos using FairMOT instead of using the benchmark's manual annotations.

Since the I2V models do not perfectly follow camera motion prompts, some of the generated videos deviate from a stationary viewpoint. We automatically filter for fixed perspective videos using Lucas-Kanade (LK) optical flow \cite{lucas1981iterative}. Our implementation samples two frames per second and uses Shi-Tomasi corner detection \cite{shi_tomasi_features} to identify distinctive features in each reference frame. These features are then tracked to subsequent frames using pyramidal LK optical flow \cite{bouguet2001pyramidal}.
A video is labeled as ``moving'' if 85\% or more of the tracked features show significant displacement (greater than 10 pixels), while videos below this threshold are classified as ``static'' with a fixed camera perspective.

\section{Results}

\begin{figure}[t]
    \centering
    \includegraphics[width=\linewidth]{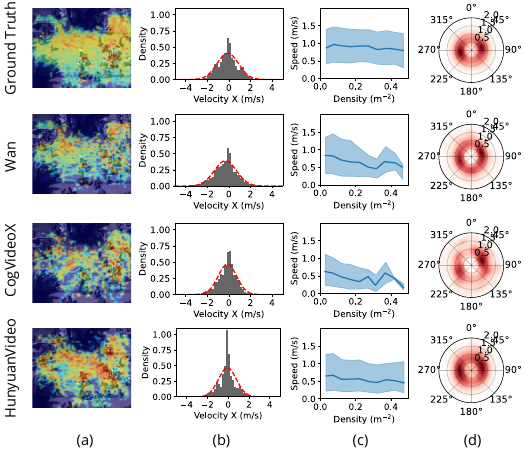} 
    \caption{Pedestrian dynamics evaluated on the UNIV scene. (a) position heatmaps; (b) velocity distribution with gaussian fit; (c) fundamental diagram; and (d) nearest neighbor polar heatmap.}
	\label{fig:results}
\end{figure}

Qualitatively, all models tested produce videos that appear natural at first glance (Fig. \ref{fig:frames}). However, closer inspection reveals anomalies that are physically implausible. For example, there is visual distortion at the bottom of the frame in the CogVideoX generation.
The models produce frame-to-frame inconsistencies with vanishing pedestrians. Figure \ref{fig:frames} shows that pedestrian ID 6 from HunyuanVideo disappears between the first and second visualized frames (0s to 1.25s), and pedestrian ID 32 in the CogVideoX sequence vanishes between the second and third images (1.25s to 2.5s).

All models achieve realistic spatial distributions of $(x,y)$ coordinates (Fig. \ref{fig:results}a), which is expected since the generations were conditioned on images with ground-truth starting positions. 
Nevertheless, this reveals that the simulated pedestrians follow some degree of scene obstacle avoidance and primarily follow sidewalks.
The heatmap from Wan2.1 generations most closely matches the ground truth.

Next, we detail the quantitative pedestrian dynamics metrics inspired by recent work in crowd simulation \cite{minartz_necs_2025,bae_continuous_2025}.

\textbf{Longitudinal velocity distribution}: We consider the velocities of non-stationary pedestrians along the primary direction of motion in the scene (Fig. \ref{fig:results}b). The ground truth distribution is roughly Gaussian but with local deviations, including elevated density at low velocities. Wan2.1 and CogVideoX reproduce the overall shape of the ground truth but with slight variations, such as heavier left tails (faster left-moving pedestrians). HunyuanVideo concentrates more samples at low velocities compared to the ground truth.

\textbf{Fundamental Diagram}: The fundamental diagram (Fig. \ref{fig:results}c) is a classical measure of crowd dynamics that shows the relationship between crowd density and flow velocity \cite{Seyfried_2005}. The local crowd density around a pedestrian is computed using Voronoi tesselation. It is typically expected to show decreasing speed with increasing crowd density, which is observed subtly in the ground truth data and more prominiently in the simulated video.

\textbf{Nearest neighbor distribution}: We include polar-coordinate density plots showing the distribution of the relative (agent-centric) positions of each pedestrian's nearest neighbor in video frames where more than one pedestrian is detected. Prior research has noted that this position typically follows a bimodal distribution with peaks at distances around 0.5-0.75 meters \cite{minartz_necs_2025}. The ground truth data indeed follows this pattern. Most of the I2V models capture roughly the same distribution (Fig. \ref{fig:results}d). CogVideoX produces the noisiest distribution, while HunyuanVideo most closely matches the ground truth.

\textbf{Statistics}: We also include a number of tabulated dynamics statistics in Table \ref{tab:results}. First, we count the proportion of pedestrians in the scene which are stationary, which is defined as pedestrians that move less than 0.2 meters from start to finish. The average speed is the mean velocity magnitude in meters/sec.
The distance traveled is the total path length, or the cumulative Euclidean distance across all positions in each  trajectory.
Finally, the passing distance is the closest distance (in meters) between two nearest-neighbor pedestrians that approach one another within 10 m.
No single model consistently matches the ground truth across all scenarios.
Wan2.1 is best at reproducing realistic walking speeds. HunyuanVideo best captures the proportions of stationary pedestrians. Passing distance proves challenging with no clear winner, suggesting that agent-to-agent interaction modeling remains a weakness.

\def\metricB{\% Stationary}
\def\metricC{Avg Speed}
\def\metricD{Dist Traveled}
\def\metricE{Passing Dist}

\def\modelA{GT}
\def\modelB{Wan}
\def\modelC{CVX}
\def\modelD{HV}

\begin{table}[htb!]
  \centering
  \caption{Pedestrian dynamics metrics across ETH/UCY scenes in the ground truth (GT) data and the models Wan2.1 (Wan), CogVideoX1.5 (CVX), HunyuanVideo-I2V (HV). The results closest to ground truth are in bold.}
  \label{tab:results}
  \footnotesize
  \begin{tabular}{@{}llcccc@{}}
    \toprule
    \textbf{Error Metric} & \textbf{Model} & \textbf{ETH} & \textbf{UNIV} & \textbf{HOTEL} & \textbf{ZARA} \\
    \midrule
    \multirow{4}{*}{\metricB} & \modelA & $37.95$ & $26.65$ & $39.05$ & $26.30$ \\ [1pt] \cline{2-6} \addlinespace[2pt]
                             & \modelB & $47.27$ & $35.03$ & $\textbf{38.29}$ & $35.02$ \\
                             & \modelC & $22.03$ & $39.94$ & $55.07$ & $36.04$ \\
                             & \modelD & $\textbf{45.59}$ & $\textbf{25.72}$ & $47.59$ & $\textbf{24.88}$ \\
    \midrule
    \multirow{4}{*}{\metricC} & \modelA & $1.29$ & $1.30$ & $1.38$ & $1.54$ \\ [1pt] \cline{2-6} \addlinespace[2pt]
                             & \modelB & $\textbf{1.09}$ & $\textbf{1.03}$ & $\textbf{1.46}$ & $1.15$ \\
                             & \modelC & $2.43$ & $1.01$ & $1.84$ & $1.96$ \\
                             & \modelD & $1.93$ & $0.89$ & $1.05$ & $\textbf{1.50}$ \\
    \midrule
    \multirow{4}{*}{\metricD} & \modelA & $1.66$ & $3.80$ & $2.66$ & $3.38$ \\ [1pt] \cline{2-6} \addlinespace[2pt]
                             & \modelB & $\textbf{1.87}$ & $2.33$ & $\textbf{1.90}$ & $2.33$ \\
                             & \modelC & $3.03$ & $2.24$ & $1.75$ & $2.19$ \\
                             & \modelD & $2.24$ & $\textbf{2.76}$ & $1.78$ & $\textbf{3.17}$ \\
    \midrule
    \multirow{4}{*}{\metricE} & \modelA & $2.49$ & $1.57$ & $3.21$ & $3.18$ \\ [1pt] \cline{2-6} \addlinespace[2pt]
                             & \modelB & $\textbf{2.61}$ & $2.46$ & $2.65$ & $2.89$ \\
                             & \modelC & $3.39$ & $2.80$ & $2.88$ & $\textbf{3.19}$ \\
                             & \modelD & $3.82$ & $\textbf{1.78}$ & $\textbf{2.90}$ & $2.65$ \\
    \bottomrule
  \end{tabular}
\end{table}

\section{Conclusion}
We have evaluated state-of-the-art image-to-video (I2V) models for their ability to simulate realistic pedestrian behavior in crowded scenes conditioned on images from the ETH/UCY trajectory prediction benchmark. Our framework revealed limitations in current models' ability to capture multi-agent interactions like passing distance, but highlight their overall visual realism, plausible walking speeds, and agent-environment interactions. A more comprehensive future assessment could include larger closed-source models like Google's Veo 3 and OpenAI's Sora, given the documented scaling benefits for video generation models.

\section*{Acknowledgments}
Thank you to Minartz et al. for their code (\href{https://github.com/kminartz/NeCS}{NeCS}) which served as a base for some of the metrics and plots. 

Financial support for the authors was provided by the U.S. ONR under grant N00014-23-1-2799.

\bibliography{references}

\begin{thebibliography}{41}
\providecommand{\natexlab}[1]{#1}
\providecommand{\url}[1]{\texttt{#1}}
\expandafter\ifx\csname urlstyle\endcsname\relax
  \providecommand{\doi}[1]{doi: #1}\else
  \providecommand{\doi}{doi: \begingroup \urlstyle{rm}\Url}\fi

\bibitem[Al-Kodmany(2013)]{al-kodmany_crowd_2013}
Al-Kodmany, K.
\newblock Crowd management and urban design: {New} scientific approaches.
\newblock \emph{URBAN DESIGN International}, 18\penalty0 (4):\penalty0 282--295, December 2013.
\newblock ISSN 1468-4519.
\newblock \doi{10.1057/udi.2013.7}.

\bibitem[Alahi et~al.(2016)Alahi, Goel, Ramanathan, Robicquet, Fei-Fei, and Savarese]{alahi_social_2016}
Alahi, A., Goel, K., Ramanathan, V., Robicquet, A., Fei-Fei, L., and Savarese, S.
\newblock {Social LSTM: Human Trajectory Prediction in Crowded Spaces}.
\newblock In \emph{2016 IEEE Conference on Computer Vision and Pattern Recognition (CVPR)}, pp.\  961--971, 2016.
\newblock \doi{10.1109/CVPR.2016.110}.

\bibitem[Armanto et~al.(2025)Armanto, Rosyid, Muladi, and Gunawan]{ARMANTO2025100875}
Armanto, H., Rosyid, H.~A., Muladi, and Gunawan.
\newblock Improved non-player character (npc) behavior using evolutionary algorithm—a systematic review.
\newblock \emph{Entertainment Computing}, 52:\penalty0 100875, 2025.
\newblock ISSN 1875-9521.
\newblock \doi{https://doi.org/10.1016/j.entcom.2024.100875}.

\bibitem[Bae et~al.(2025)Bae, Lee, and Jeon]{bae_continuous_2025}
Bae, I., Lee, J., and Jeon, H.-G.
\newblock Continuous locomotive crowd behavior generation.
\newblock In \emph{Proceedings of the IEEE/CVF Conference on Computer Vision and Pattern Recognition}, 2025.

\bibitem[Bouguet et~al.(2001)]{bouguet2001pyramidal}
Bouguet, J.-Y. et~al.
\newblock Pyramidal implementation of the affine lucas kanade feature tracker description of the algorithm.
\newblock \emph{Intel corporation}, 5\penalty0 (1-10):\penalty0 4, 2001.

\bibitem[Bradley(1993)]{bradley1993proposed}
Bradley, G.
\newblock A proposed mathematical model for computer prediction of crowd movements and their associated risks.
\newblock In \emph{Proceedings of the International Conference on Engineering for Crowd Safety}, pp.\  303--311. Elsevier Publishing Company London, 1993.

\bibitem[Brooks et~al.(2024)Brooks, Peebles, Holmes, DePue, Guo, Jing, Schnurr, Taylor, Luhman, Luhman, et~al.]{brooks_video_2024}
Brooks, T., Peebles, B., Holmes, C., DePue, W., Guo, Y., Jing, L., Schnurr, D., Taylor, J., Luhman, T., Luhman, E., et~al.
\newblock Video generation models as world simulators.
\newblock \emph{OpenAI Blog}, 1:\penalty0 8, 2024.

\bibitem[Chen et~al.(2024)Chen, Zhang, Cun, Xia, Wang, Weng, and Shan]{chen_videocrafter2_2024}
Chen, H., Zhang, Y., Cun, X., Xia, M., Wang, X., Weng, C., and Shan, Y.
\newblock Videocrafter2: Overcoming data limitations for high-quality video diffusion models.
\newblock In \emph{Proceedings of the IEEE/CVF Conference on Computer Vision and Pattern Recognition (CVPR)}, pp.\  7310--7320, June 2024.

\bibitem[Chib et~al.(2024)Chib, Nath, Kabra, Gupta, and Singh]{chib_ms-tip_2024}
Chib, P.~S., Nath, A., Kabra, P., Gupta, I., and Singh, P.
\newblock {MS-TIP: imputation aware pedestrian trajectory prediction}.
\newblock In \emph{Proceedings of the 41st International Conference on Machine Learning}, ICML'24. JMLR.org, 2024.

\bibitem[Gu et~al.(2022)Gu, Chen, Li, Lin, Rao, Zhou, and Lu]{gu_stochastic_2022}
Gu, T., Chen, G., Li, J., Lin, C., Rao, Y., Zhou, J., and Lu, J.
\newblock Stochastic trajectory prediction via motion indeterminacy diffusion.
\newblock In \emph{Proceedings of the IEEE/CVF Conference on Computer Vision and Pattern Recognition}, pp.\  17113--17122, 2022.

\bibitem[Gupta et~al.(2018)Gupta, Johnson, Fei-Fei, Savarese, and Alahi]{gupta_social_2018}
Gupta, A., Johnson, J., Fei-Fei, L., Savarese, S., and Alahi, A.
\newblock Social gan: Socially acceptable trajectories with generative adversarial networks.
\newblock In \emph{2018 IEEE/CVF Conference on Computer Vision and Pattern Recognition}, pp.\  2255--2264, 2018.
\newblock \doi{10.1109/CVPR.2018.00240}.

\bibitem[Helbing \& Moln\'ar(1995)Helbing and Moln\'ar]{helbing1995social}
Helbing, D. and Moln\'ar, P.
\newblock Social force model for pedestrian dynamics.
\newblock \emph{Phys. Rev. E}, 51:\penalty0 4282--4286, May 1995.
\newblock \doi{10.1103/PhysRevE.51.4282}.

\bibitem[Huang et~al.(2024)Huang, He, Yu, Zhang, Si, Jiang, Zhang, Wu, Jin, Chanpaisit, Wang, Chen, Wang, Lin, Qiao, and Liu]{huang_vbench_2024}
Huang, Z., He, Y., Yu, J., Zhang, F., Si, C., Jiang, Y., Zhang, Y., Wu, T., Jin, Q., Chanpaisit, N., Wang, Y., Chen, X., Wang, L., Lin, D., Qiao, Y., and Liu, Z.
\newblock Vbench: Comprehensive benchmark suite for video generative models.
\newblock In \emph{Proceedings of the IEEE/CVF Conference on Computer Vision and Pattern Recognition (CVPR)}, pp.\  21807--21818, June 2024.

\bibitem[Jahanmahin et~al.(2022)Jahanmahin, Masoud, Rickli, and Djuric]{jahanmahin_human-robot_2022}
Jahanmahin, R., Masoud, S., Rickli, J., and Djuric, A.
\newblock Human-robot interactions in manufacturing: {A} survey of human behavior modeling.
\newblock \emph{Robotics and Computer-Integrated Manufacturing}, 78:\penalty0 102404, December 2022.
\newblock ISSN 0736-5845.
\newblock \doi{10.1016/j.rcim.2022.102404}.

\bibitem[Jiang et~al.(2023)Jiang, Cornman, Park, Sapp, Zhou, and Anguelov]{jiang_motiondiffuser_2023}
Jiang, C.~M., Cornman, A., Park, C., Sapp, B., Zhou, Y., and Anguelov, D.
\newblock {MotionDiffuser}: {Controllable} {Multi}-{Agent} {Motion} {Prediction} {Using} {Diffusion}.
\newblock In \emph{2023 {IEEE}/{CVF} {Conference} on {Computer} {Vision} and {Pattern} {Recognition} ({CVPR})}, pp.\  9644--9653, Vancouver, BC, Canada, June 2023. IEEE.
\newblock \doi{10.1109/cvpr52729.2023.00930}.

\bibitem[Ju et~al.(2025)Ju, Ye, Liu, Wang, Wang, Wan, Zhang, Gai, and Xu]{ju_fulldit_2025}
Ju, X., Ye, W., Liu, Q., Wang, Q., Wang, X., Wan, P., Zhang, D., Gai, K., and Xu, Q.
\newblock Fulldit: Multi-task video generative foundation model with full attention.
\newblock \emph{CoRR}, abs/2503.19907, March 2025.

\bibitem[Kong et~al.(2025)Kong, Tian, Zhang, Min, Dai, Zhou, Xiong, Li, Wu, Zhang, Wu, Lin, Yuan, Long, Wang, Wang, Li, Huang, Yang, Tan, Wang, Song, Bai, Wu, Xue, Wang, Wang, Liu, Li, Li, Wang, Yu, Deng, Li, Chen, Cui, Peng, Yu, He, Xu, Zhou, Xu, Tao, Lu, Liu, Zhou, Wang, Yang, Wang, Liu, Jiang, and Zhong]{kong_hunyuanvideo_2025}
Kong, W., Tian, Q., Zhang, Z., Min, R., Dai, Z., Zhou, J., Xiong, J., Li, X., Wu, B., Zhang, J., Wu, K., Lin, Q., Yuan, J., Long, Y., Wang, A., Wang, A., Li, C., Huang, D., Yang, F., Tan, H., Wang, H., Song, J., Bai, J., Wu, J., Xue, J., Wang, J., Wang, K., Liu, M., Li, P., Li, S., Wang, W., Yu, W., Deng, X., Li, Y., Chen, Y., Cui, Y., Peng, Y., Yu, Z., He, Z., Xu, Z., Zhou, Z., Xu, Z., Tao, Y., Lu, Q., Liu, S., Zhou, D., Wang, H., Yang, Y., Wang, D., Liu, Y., Jiang, J., and Zhong, C.
\newblock {HunyuanVideo}: {A} {Systematic} {Framework} {For} {Large} {Video} {Generative} {Models}, March 2025.

\bibitem[Lerner et~al.(2007)Lerner, Chrysanthou, and Lischinski]{Lerner_ETH}
Lerner, A., Chrysanthou, Y., and Lischinski, D.
\newblock Crowds by example.
\newblock \emph{Computer Graphics Forum}, 26\penalty0 (3):\penalty0 655--664, 2007.
\newblock \doi{https://doi.org/10.1111/j.1467-8659.2007.01089.x}.

\bibitem[Liu et~al.(2025)Liu, Ren, Gupta, and Wang]{leonardis_physgen_2025}
Liu, S., Ren, Z., Gupta, S., and Wang, S.
\newblock {PhysGen}: {Rigid}-{Body} {Physics}-{Grounded} {Image}-to-{Video} {Generation}.
\newblock In Leonardis, A., Ricci, E., Roth, S., Russakovsky, O., Sattler, T., and Varol, G. (eds.), \emph{Computer {Vision} – {ECCV} 2024}, volume 15140, pp.\  360--378. Springer Nature Switzerland, Cham, 2025.
\newblock ISBN 978-3-031-73006-1 978-3-031-73007-8.
\newblock \doi{10.1007/978-3-031-73007-8_21}.

\bibitem[Lucas \& Kanade(1981)Lucas and Kanade]{lucas1981iterative}
Lucas, B.~D. and Kanade, T.
\newblock An iterative image registration technique with an application to stereo vision.
\newblock In \emph{IJCAI'81: 7th international joint conference on Artificial intelligence}, volume~2, pp.\  674--679, 1981.

\bibitem[Minartz et~al.(2025)Minartz, Hendriks, Koop, Corbetta, and Menkovski]{minartz_necs_2025}
Minartz, K., Hendriks, F., Koop, S.~M., Corbetta, A., and Menkovski, V.
\newblock Discovering interaction mechanisms in crowds via deep generative surrogate experiments.
\newblock \emph{Scientific Reports}, 15\penalty0 (1):\penalty0 10385, March 2025.
\newblock ISSN 2045-2322.
\newblock \doi{10.1038/s41598-025-92566-9}.

\bibitem[Mohamed et~al.(2020)Mohamed, Qian, Elhoseiny, and Claudel]{mohamed_social-stgcnn_2020}
Mohamed, A., Qian, K., Elhoseiny, M., and Claudel, C.
\newblock {Social-STGCNN: A Social Spatio-Temporal Graph Convolutional Neural Network for Human Trajectory Prediction}.
\newblock In \emph{2020 IEEE/CVF Conference on Computer Vision and Pattern Recognition (CVPR)}, pp.\  14412--14420, 2020.
\newblock \doi{10.1109/CVPR42600.2020.01443}.

\bibitem[Montanaro et~al.(2024)Montanaro, Savant~Aira, Aiello, Valsesia, and Magli]{montanaro_motioncraft_nodate}
Montanaro, A., Savant~Aira, L., Aiello, E., Valsesia, D., and Magli, E.
\newblock {MotionCraft: Physics-Based Zero-Shot Video Generation}.
\newblock In Globerson, A., Mackey, L., Belgrave, D., Fan, A., Paquet, U., Tomczak, J., and Zhang, C. (eds.), \emph{Advances in Neural Information Processing Systems}, volume~37, pp.\  123155--123181. Curran Associates, Inc., 2024.

\bibitem[Nocentini et~al.(2019)Nocentini, Fiorini, Acerbi, Sorrentino, Mancioppi, and Cavallo]{nocentini_survey_2019}
Nocentini, O., Fiorini, L., Acerbi, G., Sorrentino, A., Mancioppi, G., and Cavallo, F.
\newblock A {Survey} of {Behavioral} {Models} for {Social} {Robots}.
\newblock \emph{Robotics}, 8\penalty0 (3):\penalty0 54, September 2019.
\newblock ISSN 2218-6581.
\newblock \doi{10.3390/robotics8030054}.

\bibitem[Pellegrini et~al.(2009)Pellegrini, Ess, Schindler, and van Gool]{Pellegrini_UCY}
Pellegrini, S., Ess, A., Schindler, K., and van Gool, L.
\newblock You'll never walk alone: Modeling social behavior for multi-target tracking.
\newblock In \emph{2009 IEEE 12th International Conference on Computer Vision}, pp.\  261--268, 2009.
\newblock \doi{10.1109/ICCV.2009.5459260}.

\bibitem[Reynolds et~al.(1999)]{reynolds1999steering}
Reynolds, C.~W. et~al.
\newblock Steering behaviors for autonomous characters.
\newblock In \emph{Game developers conference}, volume 1999, pp.\  763--782. Citeseer, 1999.

\bibitem[Ribeiro-Gomes et~al.(2024)Ribeiro-Gomes, Cai, Milacski, Wu, Prakash, Takagi, Aubel, Kim, Bernardino, and De~La~Torre]{ribeiro-gomes_motiongpt_2024}
Ribeiro-Gomes, J., Cai, T., Milacski, Z.~A., Wu, C., Prakash, A., Takagi, S., Aubel, A., Kim, D., Bernardino, A., and De~La~Torre, F.
\newblock {MotionGPT}: {Human} {Motion} {Synthesis} with {Improved} {Diversity} and {Realism} via {GPT}-3 {Prompting}.
\newblock In \emph{2024 {IEEE}/{CVF} {Winter} {Conference} on {Applications} of {Computer} {Vision} ({WACV})}, pp.\  5058--5068, Waikoloa, HI, USA, January 2024. IEEE.
\newblock ISBN 9798350318920.
\newblock \doi{10.1109/WACV57701.2024.00499}.

\bibitem[Salzmann et~al.(2020)Salzmann, Ivanovic, Chakravarty, and Pavone]{salzmann_trajectron_2020}
Salzmann, T., Ivanovic, B., Chakravarty, P., and Pavone, M.
\newblock Trajectron++: {Dynamically}-{Feasible} {Trajectory} {Forecasting} with {Heterogeneous} {Data}.
\newblock In Vedaldi, A., Bischof, H., Brox, T., and Frahm, J.-M. (eds.), \emph{Computer {Vision} – {ECCV} 2020}, pp.\  683--700, Cham, 2020. Springer International Publishing.
\newblock ISBN 978-3-030-58523-5.
\newblock \doi{10.1007/978-3-030-58523-5_40}.

\bibitem[Seyfried et~al.(2005)Seyfried, Steffen, Klingsch, and Boltes]{Seyfried_2005}
Seyfried, A., Steffen, B., Klingsch, W., and Boltes, M.
\newblock The fundamental diagram of pedestrian movement revisited.
\newblock \emph{Journal of Statistical Mechanics: Theory and Experiment}, 2005\penalty0 (10):\penalty0 P10002, oct 2005.
\newblock \doi{10.1088/1742-5468/2005/10/P10002}.

\bibitem[Shi \& Tomasi(1994)Shi and Tomasi]{shi_tomasi_features}
Shi, J. and Tomasi.
\newblock Good features to track.
\newblock In \emph{1994 Proceedings of IEEE Conference on Computer Vision and Pattern Recognition}, pp.\  593--600, 1994.
\newblock \doi{10.1109/CVPR.1994.323794}.

\bibitem[Shi et~al.(2021)Shi, Wang, Long, Zhou, Zhou, Niu, and Hua]{shi_sgcn_2021}
Shi, L., Wang, L., Long, C., Zhou, S., Zhou, M., Niu, Z., and Hua, G.
\newblock {SGCN: Sparse Graph Convolution Network for Pedestrian Trajectory Prediction}.
\newblock In \emph{2021 IEEE/CVF Conference on Computer Vision and Pattern Recognition (CVPR)}, pp.\  8990--8999, 2021.
\newblock \doi{10.1109/CVPR46437.2021.00888}.

\bibitem[van~den Berg et~al.(2008)van~den Berg, Lin, and Manocha]{van2008reciprocal}
van~den Berg, J., Lin, M., and Manocha, D.
\newblock {Reciprocal Velocity Obstacles for real-time multi-agent navigation}.
\newblock In \emph{2008 IEEE International Conference on Robotics and Automation}, pp.\  1928--1935, 2008.
\newblock \doi{10.1109/ROBOT.2008.4543489}.

\bibitem[Wan et~al.(2025)Wan, Wang, Ai, Wen, Mao, Xie, Chen, Yu, Zhao, Yang, Zeng, Wang, Zhang, Zhou, Wang, Chen, Zhu, Zhao, Yan, Huang, Feng, Zhang, Li, Wu, Chu, Feng, Zhang, Sun, Fang, Wang, Gui, Weng, Shen, Lin, Wang, Wang, Zhou, Wang, Shen, Yu, Shi, Huang, Xu, Kou, Lv, Li, Liu, Wang, Zhang, Huang, Li, Wu, Liu, Pan, Zheng, Hong, Shi, Feng, Jiang, Han, Wu, and Liu]{wan_wan_2025}
Wan, T., Wang, A., Ai, B., Wen, B., Mao, C., Xie, C.-W., Chen, D., Yu, F., Zhao, H., Yang, J., Zeng, J., Wang, J., Zhang, J., Zhou, J., Wang, J., Chen, J., Zhu, K., Zhao, K., Yan, K., Huang, L., Feng, M., Zhang, N., Li, P., Wu, P., Chu, R., Feng, R., Zhang, S., Sun, S., Fang, T., Wang, T., Gui, T., Weng, T., Shen, T., Lin, W., Wang, W., Wang, W., Zhou, W., Wang, W., Shen, W., Yu, W., Shi, X., Huang, X., Xu, X., Kou, Y., Lv, Y., Li, Y., Liu, Y., Wang, Y., Zhang, Y., Huang, Y., Li, Y., Wu, Y., Liu, Y., Pan, Y., Zheng, Y., Hong, Y., Shi, Y., Feng, Y., Jiang, Z., Han, Z., Wu, Z.-F., and Liu, Z.
\newblock {Wan: Open and Advanced Large-Scale Video Generative Models}.
\newblock \emph{arXiv preprint arXiv:2503.20314}, 2025.

\bibitem[Yang et~al.(2024)Yang, Teng, Zheng, Ding, Huang, Xu, Yang, Hong, Zhang, Feng, et~al.]{yang_cogvideox_2025}
Yang, Z., Teng, J., Zheng, W., Ding, M., Huang, S., Xu, J., Yang, Y., Hong, W., Zhang, X., Feng, G., et~al.
\newblock {CogVideoX: Text-to-Video Diffusion Models with An Expert Transformer}.
\newblock \emph{arXiv preprint arXiv:2408.06072}, 2024.

\bibitem[Yao et~al.(2024)Yao, Zhu, Bi, Mao, and Wang]{trajclip_2024}
Yao, P., Zhu, Y., Bi, H., Mao, T., and Wang, Z.
\newblock Trajclip: Pedestrian trajectory prediction method using contrastive learning and idempotent networks.
\newblock In Globerson, A., Mackey, L., Belgrave, D., Fan, A., Paquet, U., Tomczak, J., and Zhang, C. (eds.), \emph{Advances in Neural Information Processing Systems}, volume~37, pp.\  77023--77037. Curran Associates, Inc., 2024.

\bibitem[Yuan et~al.(2021)Yuan, Weng, Ou, and Kitani]{yuan_agentformer_2021}
Yuan, Y., Weng, X., Ou, Y., and Kitani, K.
\newblock {AgentFormer}: {Agent}-{Aware} {Transformers} for {Socio}-{Temporal} {Multi}-{Agent} {Forecasting}.
\newblock In \emph{2021 {IEEE}/{CVF} {International} {Conference} on {Computer} {Vision} ({ICCV})}, pp.\  9793--9803, Montreal, QC, Canada, October 2021. IEEE.
\newblock ISBN 978-1-66542-812-5.
\newblock \doi{10.1109/ICCV48922.2021.00967}.

\bibitem[Zhang et~al.(2024)Zhang, Yu, Wu, Feng, Zheng, Snavely, Wu, and Freeman]{zhang_physics-based_nodate}
Zhang, T., Yu, H.-X., Wu, R., Feng, B.~Y., Zheng, C., Snavely, N., Wu, J., and Freeman, W.~T.
\newblock Physics-based interaction with 3d objects via video generation.
\newblock In \emph{Proceedings of the European conference on computer vision (ECCV)}, volume~2, 2024.

\bibitem[Zhang et~al.(2021)Zhang, Wang, Wang, Zeng, and Liu]{zhang_fairmot_2021}
Zhang, Y., Wang, C., Wang, X., Zeng, W., and Liu, W.
\newblock {FairMOT}: {On} the {Fairness} of {Detection} and {Re}-identification in {Multiple} {Object} {Tracking}.
\newblock \emph{International Journal of Computer Vision}, 129\penalty0 (11):\penalty0 3069--3087, November 2021.
\newblock ISSN 1573-1405.
\newblock \doi{10.1007/s11263-021-01513-4}.

\bibitem[Zheng et~al.(2025)Zheng, Huang, Liu, Zou, He, Zhang, Zhang, He, Zheng, Qiao, and Liu]{zheng_vbench-20_2025}
Zheng, D., Huang, Z., Liu, H., Zou, K., He, Y., Zhang, F., Zhang, Y., He, J., Zheng, W.-S., Qiao, Y., and Liu, Z.
\newblock {VBench}-2.0: {Advancing} {Video} {Generation} {Benchmark} {Suite} for {Intrinsic} {Faithfulness}.
\newblock \emph{arXiv preprint arXiv:2503.21755}, 2025.

\bibitem[Zheng et~al.(2024)Zheng, Peng, Yang, Shen, Li, Liu, Zhou, Li, and You]{zheng2024open}
Zheng, Z., Peng, X., Yang, T., Shen, C., Li, S., Liu, H., Zhou, Y., Li, T., and You, Y.
\newblock Open-sora: Democratizing efficient video production for all.
\newblock \emph{arXiv preprint arXiv:2412.20404}, 2024.

\bibitem[Zhou et~al.(2010)Zhou, Chen, Cai, Luo, Low, Tian, Tay, Ong, and Hamilton]{zhou_crowd_2010}
Zhou, S., Chen, D., Cai, W., Luo, L., Low, M. Y.~H., Tian, F., Tay, V. S.-H., Ong, D. W.~S., and Hamilton, B.~D.
\newblock Crowd modeling and simulation technologies.
\newblock \emph{ACM Transactions on Modeling and Computer Simulation}, 20\penalty0 (4):\penalty0 1--35, October 2010.
\newblock ISSN 1049-3301, 1558-1195.
\newblock \doi{10.1145/1842722.1842725}.

\end{thebibliography}
\bibliographystyle{icml2025}

\end{document}